# Modular, Hierarchical Machine Learning for Sequential Goal Completion

Nathan R. McDonald*[a]

[a]Air Force Research Laboratory, Information Directorate, 525 Brooks Road, Rome, NY 13441, USA

## ABSTRACT

Given a maze populated with different objects, one may task a robot with a sequential goal completion task, e.g. 1) pick up a key then 2) unlock the door then 3) unlock the treasure chest. A typical machine learning (ML) solution would involve a monolithically trained artificial neural network (ANN). However, if the sequence of goals or the goals themselves change, then the ANN must be significantly (or, at worst, completely) retrained. Instead of a monolithic ANN, a modular ML component would be 1) independently optimizable (task-agnostic) and 2) arbitrarily reconfigurable with other ML modules. This work describes a modular, hierarchical ML framework by integrating two emerging ML techniques: 1) cognitive map learners (CML) and 2) hyperdimensional computing (HDC). A CML is a collection of three single layer ANNs (matrices) collaboratively trained to learn the topology of an abstract graph. Here, two CMLs were constructed, one describing locations on in 2D physical space and the other the relative distribution of objects found in this space. Each CML node states was encoded as a high-dimensional vector to utilize HDC, an ML algebra, for symbolic reasoning over these high-dimensional "symbol" vectors. In this way, each sub-goal above was described by algebraic equations of CML node states. Multiple, independently trained CMLs were subsequently assembled together to navigate a maze to solve a sequential goal task. Critically, changes to these goals required only localized changes in the CML-HDC architecture, as opposed to a global ANN retraining scheme. This framework therefore enabled a more traditional engineering approach to ML, akin to digital logic design.

**Keywords:** hyperdimensional computing, vector symbolic architectures, cognitive map learners, artificial neural networks, neuroengineering, path planning, modular machine learning

## 1. INTRODUCTION

Since deep neural networks (DNN) are typically trained monolithically, or end-to-end, to solve a well-defined task, sequential goal completion tasks are difficult. With respect to maze puzzles, DNNs struggle with learning from minimal training data, knowledge transfer, generalization to novel environments, and generating human-interpretable models [1]. From a classical mathematics perspective, many planning tasks can be formulated as finding the shortest path on an abstract graph [2]. However, the standard Dijkstra and A* shortest path algorithms must compute the entire route before deciding the first step. If the goal location changes before the algorithms finishes, then the whole algorithm must be restarted from scratch. Such path planners are then not ideal for autonomous, expeditionary robotics, which must rapidly respond to dynamic environments.

Alternatively, a modular machine learning (ML) approach could segregate knowledge in modules, e.g. movement, spatial relationships, and positions. By encoding this information and decision space in a consistent information representation, multiple neural network modules can be independently prepared (learned or calculated) and integrated together into a larger assembly, in a manner similar to digital logic. This work demonstrated a modular, hierarchical ML framework implemented according to two emerging ML techniques: 1) cognitive map learners and 2) hyperdimensional computing.

Cognitive map learners (CML) are a new approach to artificial neural networks (ANN) and are trained to learn the topology of an abstract graph [3]. The CML's three separate yet collaboratively trained single-layer ANNs (matrices) each learn internal representations of a different aspect of the graph: 1) node states, 2) edge actions, and 3) edge action availability. As a result of this atypical segregation of information, the CML, though never explicitly trained for path planning, can iteratively compute a near optimal path (fewest edges) between any initial and target node state [3].

However, the CML has no mechanism for self-selecting a target node state; rather, an external source must specify a target state to begin the CML computation. Compartmentalization of information in a CML permits post-training "brain surgery" to extract these internal state representations. Hyperdimensional computing (HDC), or Vector Symbolic Architectures (VSA), [4, 5], is a mathematical algebra well suited for integrating and orchestrating multiple CMLs together. Instead of learning synaptic weights values, HDC encodes learning by manipulating the similarity among a set of high-dimensional vectors [6, 7, 8]. Being an algebra, such learning is explicitly expressed in equations, affording both human interpretation and intervention [9]. HDC-based CMLs can receive external inputs and compute output responses which are semantically meaningful for other HDC-based modules.

*nathan.mcdonald.5@us.af.mil

This work considers a path planning task through a maze along an arbitrary sequence of objects (Fig. 1a). While the number and relative location of the eight objects are consistent, the precise locations changed per trial. One CML learned an abstract graph describing the relative positions of these objects (Fig. 1b), while a second CML learned to navigate a 2D Cartesian grid (Fig. 4a). By integrating the object CML, grid CML, and robot touch sensor data together via HDC, a simulated robot successfully navigated among an arbitrary sequence of objects in a maze along a near optimal path in all trials. This work suggests a template for building hierarchies of biologically plausible cognitive abstraction and orchestration.

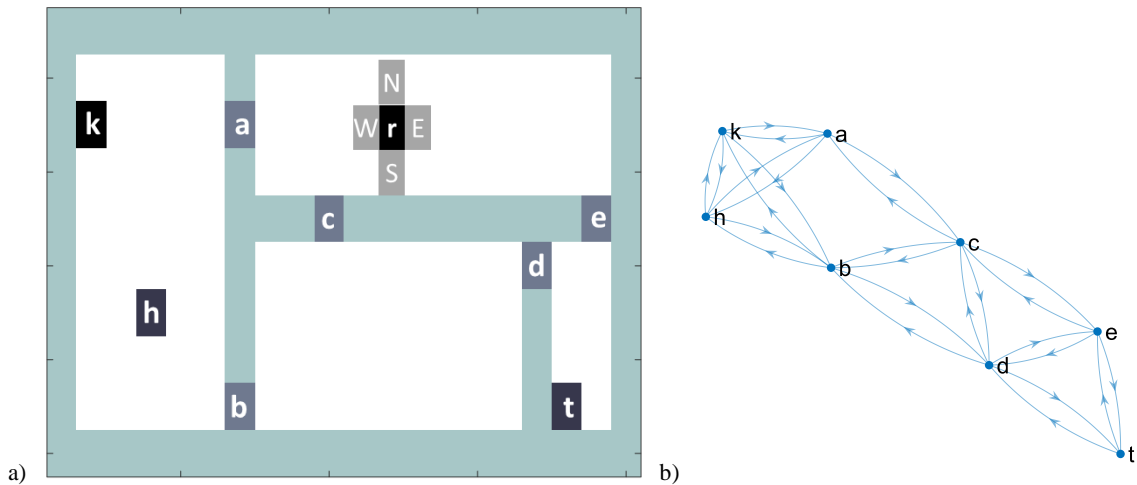

Figure 1. a) Sample maze, where the robot, $r$, must navigate from home, $h$, to the key, $k$, to the treasure, $t$. The robot may only move through the five doors, $a$, $b$, $c$, $d$, and $e$, so four touch sensors, $N$, $E$, $S$, $W$, were provided to negotiate the variable wall locations (green). b) Abstract graph used for the object CML $C_O$, where nodes represent the 8 maze objects and edges indicate line of sight path.

The contributions of the work are as follows:

1) Templated decomposition of variable maze problem into several smaller sub-tasks, viz. relative object positioning, 2D spatial navigation, and trial-specific object positions.

2) Integrated each of the aforesaid modular ML solutions into a hierarchical ML solution via HDC algebra

3) Demonstrated neural network based real-time path planning and sequential goal completion.

With respect to mathematical notation, matrices are denoted by capital letters and vectors by lower case letters. Importantly, lower case letter vectors come from matrices of the same uppercase letter, e.g. $s_i$ denotes the $i^{th}$ row/column vector of matrix $S$. Key symbols are consolidated and defined in Table I. (Appendix)

## 2. BACKGROUND

### 2.1 Cognitive map learner (CML)

A cognitive map learner is a collection of three single layer neural networks (matrices) which are trained or calculated to encode the topology of a graph of $n$ nodes and $e$ edges. Each neural network matrix learns one aspect of the graph, 1) node state representations $S \in \mathbb{R}^{(d,n)}$, 2) edge action representations $A \in \mathbb{R}^{(d,e)}$, and 3) the availability $G \in \mathbb{R}^{(e,n)}$ of edge actions from each node state, where $d$ is the vector length. By segregating knowledge in this way, the CML as a modular unit can iteratively plan the near-optimal shortest path between any two nodes in the graph (Fig. 2a) [3]. This solution is competitive with Dijkstra and A* [2], though it lacks the same mathematical optimality guarantees [3]. Unlike Dijkstra and A*, however, CMLs provides partial solutions in real-time, rerouting to accommodate for changing target nodes or dropped edges.

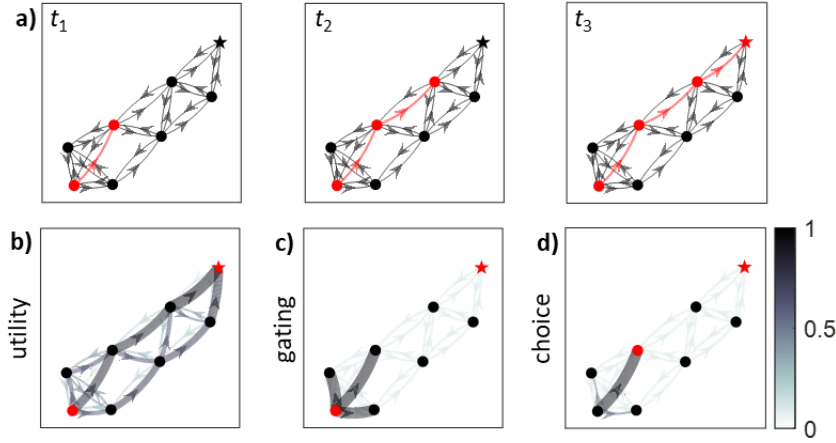

Figure 2. a) CML iteratively path planning (red edges) from its current state (red circle) to the target state (star). Detailing the first path choice $t_1$ further, b) the edge action utility values $u$ and c) the gating values $g$ are multiplied together as d) $g \odot u$ to determine the choice of new current node state (red circle). Edge color and width scale with the color bar for visual clarity.

To train a CML on an abstract graph, the node state matrix $S$ and edge action matrix $A$ are initialized with random Gaussian values, $\mu = 0$, with $\sigma = 0.1$ and $\sigma = 1$, respectively. The gating matrix $G$ encodes the permissibility values of each edge action from every node. For an unweighted graph, permissible actions have values of 1 and 0 otherwise; for weighted graphs, the weight $w$ is encoded as the reciprocal, $1/w$.

The following is a simplified version of the original CML algorithm framework [3] to provide clearer access to the matrices $S$, $A$, and $G$. During training, the network is tasked to predict ( or approximate) the next node state $s_{t+1}$ given its current node state $s_t$ and choice of edge action $a_t$.

$$s_{t+1} = s_t + a_t. \tag{1}$$

Each column of $S$ and $A$ are updated according to a bio-plausible delta learning rule [10], which modifies the weights as the difference between the actual $s_{t+1}$ and the predicted $\hat{s}_{t+1}$ values,

$$\Delta a_t = l\,(s_{t+1} - \hat{s}_{t+1}), \tag{2}$$

$$\Delta s_t = l\,(\hat{s}_{t+1} - s_{t+1}), \tag{3}$$

where $l$ is the learning rate. For simplicity, weight updates are summed and applied at the end of each training epoch, which is all $e$ actions. Training continues until the system converges. Note, as will become important later, learned node state vectors are pseudo-orthogonal to each other [11]. Thus, alternatively, one can initialize $S$ with pseudo-orthogonal node state vectors and calculate each $A$ directly as

$$a_{ij} = s_i - s_j. \tag{4}$$

By learning these pairwise node associations, the CML learns the topology of the graph. The CML itself has no mechanism for deciding when or why to begin plotting a path from its current node state $s_t$ to a target node state $s^*$. Target node selection must be generated externally then provided to the CML. Given then a target node state $s^*$, at each time step, the utility of every edge action with respect to $s^*$ is calculated by multiplying the difference between the target and current node state by the pseudo-inverse of edge action matrix $A$ [3] (Fig. 2b),

$$u = A^\dagger\,(s^* - s_t), \tag{5}$$

where $\dagger$ denotes the Moore-Penrose pseudo-inverse. The gating vector $g_t$ corresponding to node $s_t$ (Fig. 2c) is multiplied elementwise by the utility vector, ensuring only legal edge actions to have a nonzero utility value. A winner-take-all (*WTA*) algorithm produces a one-hot vector indicating the estimated most useful edge action (Fig. 2d),

$$\hat{s}_{t+1} = s_t + A \cdot WTA(g_t \odot u), \tag{6}$$

where $\odot$ denotes elementwise multiplication. Iterating over Eq. 5, 6), the CML finds a "reasonable" minimal path between any initial and target state.

## 2.2 Hyperdimensional computing (HDC)

CML vector length $d$ may be as small as 80 [3], but by increasing this value to $d \geq 512$, node and edge vectors become semantically meaningful entities for hyperdimensional computing algebra [4]. As opposed to artificial neurons and synapses, HDC performs symbolic reasoning with high dimensional vectors, or hypervectors. According to the law of large numbers, as the length of these randomly generated hypervectors increases, the similarity between any two converges to pseudo-orthogonality [5, 6].

Thus, if two symbol hypervectors are found to be not pseudo-orthogonal, then statistically there must be some correlation between them. This dimensionality and similarity relationship is observed for a variety of vector elements, with common implementations being binary $\{0, 1\}$ [6], bipolar $\{-1, +1\}$ [7], and complex valued $e^{j[-\pi,+\pi]}$ [8].

The following examples assume a dictionary $D$ of known hypervectors, $D = \{w, x, y, z\}$, and a random hypervector $\eta$. The cosine similarity metric measures the angle between vectors, irrespective of vector magnitude,

$$\delta(x, y) = \frac{x \cdot y}{\|x\| \|y\|}, \tag{7}$$

such that identical vectors have a cosine similarity of 1 and pseudo-orthogonal hypervectors have a similarity close to 0.

The basic operations of HDC are addition, multiplication, permutation, and recovery. Addition and multiplication are elementwise operations, so the dimension of the resultant hypervector remains $d$ regardless of the number of hypervectors added or multiplied together. Addition (or "bundling") is performed as signed addition,

$$q = sgn([x + y + z]), \tag{8}$$

where

$$sgn(x) := \begin{cases} -1 & if\ x < 0 \\ 0 & if\ x = 0 \\ 1 & if\ x > 0 \end{cases}. \tag{9}$$

When adding an even number of hypervectors, a random hypervector $\eta$ is included to break ties. The bundled hypervector is similar to each of the vectors added together, $\delta(q,x) \sim \delta(q,y) \sim \delta(q,z) \gg \delta(q,\eta) \sim 0$.

Multiplication, denoted as $\odot$, binds hypervectors together, analogous to key-value pairing. Unlike bundling, the product hypervector is not similar to either of its component hypervectors; but the operation is reversible. Given $q = [w \odot x + y \odot z]$, to approximate the hypervector bound with $w$,

$$\begin{aligned} w \odot q &= w \odot [w \odot x + y \odot z] \\ &= \cancel{w \odot w} \odot x + w \odot y \odot z \\ &= x + \eta \\ &\sim x \end{aligned} \tag{10}$$

where extraneous terms, being pseudo-orthogonal, are consolidated as a random hypervector $\eta$.

The recovery, or cleanup, operation compares a query hypervector to all other known hypervectors stored in a dictionary and returns the most similar hypervector above a minimum similarity $\theta$,

$$rec(\sim x, D, \theta) = x. \tag{11}$$

Permutation, denoted $\rho$, is simply a circular shift of the hypervector elements. This operator is also reversible, so it is often used to encode sequences. Let $q = [x + \rho(y) + \rho_2(z)]$, then

$$\rho_{-1}(q) \sim y. \tag{12}$$

# 3. METHODS

For this work, a simulated robot was tasked to autonomously navigate a variable maze according to an arbitrary sequence of objects (Fig. 1). The maze spanned a 2D 10×20 Cartesian grid. Each maze was partitioned by walls and populated with 8 objects, viz. home $h$, key $k$, treasure $t$, and doors $a, b, c, d,$ and $e$. The robot could move in the 4 cardinal directions, viz. north, south, east, and west. To prevent the robot from driving into a wall, 4 touch sensors were used, corresponding to these same directions.

To solve this sequential goal completion task in a variable maze using CMLs, the problem was decomposed into several discrete sub-components which were then assembled together via HDC. First, the sequence of goals was specified as a simple HDC equation (Sect. 3.1), which could be later updated. Next, the CML framework was prepared as a modular ML unit $C$, with defined inputs and outputs (Sect. 3.2). Next an object CML $C_O$ was prepared to learn the relative spatial relationships among the eight objects in the maze, while a grid position CML $C_P$ was trained to navigate a 10×20 grid (Sect. 3.3). Next, a *map* hypervector was created to store each object and its grid position (Sect. 3.4). Finally, all these components were assembled together in a hierarchical framework to autonomously, sequentially find the prescribed objects in various mazes (Sect. 3.5).

## 3.1 Maze

The simulated robot always started at the home position $h$ and was tasked 1) to move to the key $k$, then 2) to the treasure $t$, then 3) back to home $h$. The permute operator encoded the sequence of goals to be completed as a behavioral *policy*,

$$policy = [\rho_1(k) + \rho_2(t) + \rho_3(h)]. \tag{13}$$

This policy was provided to the robot as an input, and the robot executed the instructions. As the robot completed each subgoal, the *policy* hypervector was unpermuted by 1 to reveal the next goal.

## 3.2 CML as modular ML unit

While the CML does not have any mechanism to specify a target state itself, due to its modular construction, each learned node state may be extracted column-wise from the node state matrix $S$ and later supplied to the CML by external sources. Defined as a modular ML unit, a CML takes two inputs: the current node state $s_t$ and the target node state $s^*$. After performing a single iteration of Eq. 5, 6), the CML unit returns two outputs: an action vector $a_t$ and the next predicted (expected, or prescribed) node state $\hat{s}_{t+1}$ (Fig. 3a, Algorithm A.1 in Appendix),

$$a_t, \hat{s}_{t+1} \leftarrow C(s^*, s_t). \tag{14}$$

To plan a complete path, one lets the output $\hat{s}_{t+1}$ feedback directly to the CML input $s_t$ (Fig. 3b); and the CML runs recurrently until the current and target node state are sufficiently similar, $\delta(s^*, s_t) \geq \phi$. Additionally, while the CML will correctly plan paths even when the target vector $s^*$ is only approximately similar to an actual node state vector, e.g. $C(\sim s^*, s_t)$ [11], to prevent the CML from responding to random noise as an input, a recovery operation was included here to sanitize both the current and target node states. In Fig. 3b, this is illustrated as $s^*$ ($s_t$) becoming $s_j$ ($s_i$). If either $s^*$ or $s_t$ were not in $S$, then the CML returned a zeros vector (not shown).

For all CMLs used in the following experiments, the hypervector size was $d = 1000$. The mean similarity among random hypervectors was $\delta(s_i, s_j)_{i \neq j} = 0.00 \pm 0.031$ but could be as high as $|\delta_{max}(s_i, s_j)_{i \neq j}| < 0.1 = \theta$.

As previously mentioned, CMLs were trained until the $S$ and $A$ matrices converged or were calculated directly. Since any exhaustive verification over all node pairs could be computationally prohibitive, each CML was considered verified only if it plotted "reasonable" paths for all 50 pairs of graph nodes. Due to the expected pseudo-orthogonality of randomly generated high-dimensional vectors, viable CMLs were created from effectively any set of random vectors.

## 3.3 Object and grid position CMLs

While the exact locations of the 8 maze objects were randomly assigned each trial, their relative locations remained consistent, e.g. the key and home objects always occurred in the leftmost room (Fig. 1a). An abstract graph of the object layout was created, where each node represented an object and each edge indicated a line of sight path between objects (Fig. 1b). Each of the 8 objects was assigned one of the random bipolar node states in $O \in \mathbb{R}^{(1e3,8)}$, and the edge action vectors were calculated (Eq. 4), $A \in \mathbb{R}^{(1e3,26)}$. This object CML $C_O$ was used to determine the sequence of objects the robot would navigate to on its way to each goal in the *policy*, Eq. 13).

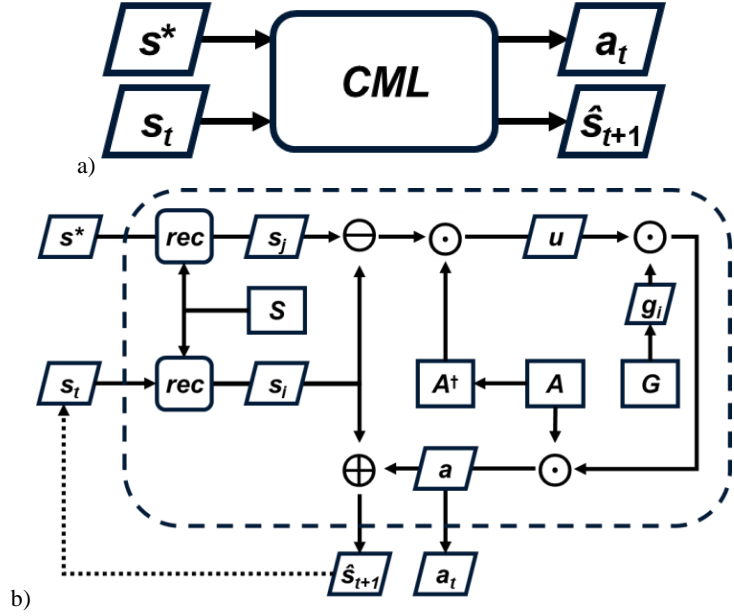

Figure 3. a) Block diagram of CML with inputs and outputs. b) Typical wiring diagram for CML computations, with optional connection between $\hat{s}_{t+1}$ and $s_t$ (dotted line). Rectangles denote static values, and parallelograms describe the information content of the hypervector along the path.

To perform the fine-grained path planning in the 2D maze, a grid position CML $C_P$ was also created. Unlike for an abstract graph CML where every edge action has a unique state representation, for graphs of physical spaces, e.g. a 2D Cartesian grid, graph symmetry can be exploited to reduce the size of the action matrix $A$ (Fig. 4a) [3]. For the 10×20 2D Cartesian maze with movement only along the cardinal directions ($e = 4$ edge actions), the node state matrix $P \in \mathbb{R}^{(1e3, 200)}$ and edge action matrix $A \in \mathbb{R}^{(1e3, 4)}$. Unlike in Eq. 4), where the CML calculated edge states in $A$ to accommodate the node states in $S$, this 2D CML fixed 4 edge actions in $A$ and learned (or calculated) the node state vectors in $P$. Two random high-dimensional vectors were generated to represent movement south and east. If moving south was defined by vector $a_s$, moving north may be defined by an anti-parallel vector $a_n$, $\delta(a_s, a_n) = -1$. Since cosine similarity measured the angle between vectors irrespective of vector magnitude, simply reversing the signs of the vector elements of south $a_s$ (east $a_e$) created the corresponding anti-parallel north $a_n$ (west $a_w$) vector. Unlike for the (abstract graph) object CML $C_O$ with pseudo-orthogonal node states $O$, the grid position CML $C_P$ node states $P$ were much more similar $\delta \sim 1$ or anti-parallel $\delta \sim -1$ with each other (Fig. 4b).

The node state matrix $P$ was initialized as a zeroes matrix and trained over all 740 graph edges (still only 4 edge action states) until convergence. While the paths through the grid space could be constructed from scratch during a robot exploration phase, for simplicity here, the space of nodes and edges for training was simply exhaustively computed. Instead of explicitly encoding walls by removing graph edges for every test maze, touch sensors were provided to the robot. In effect, these touch sensors functioned as the gating matrix $G$, prohibiting illegal actions. So for the grid position CML, the entire $G$ matrix was replaced by a vector reflecting the scalar output from each of the 4 touch sensors, where the sensor output was 0 when in contact with a wall and 1 otherwise (Fig. 5),

$$g = [E, S, N, W]^T, \quad (15)$$

where E, S, N, W represent the east, south, north, and west touch sensor value, respectively.

### 3.4 Map of objects in 2D grid

Having the object and grid position CMLs alone were insufficient to solve the maze task. A map was needed to locate the set of eight objects on the 2D grid for a particular trial. To create the map, each object node state was multiplied by the grid position node state then summed together,

$$map = sgn(\sum_i^8 sgn(o_i \odot p_i)). \quad (16)$$

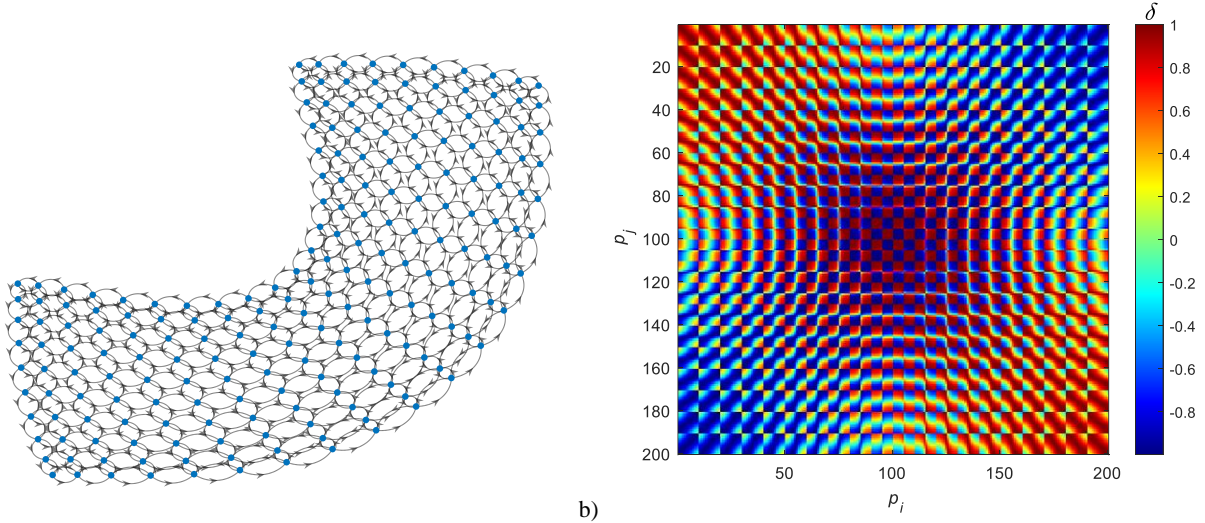

a)           b)

Figure 4. a) Graph of 10×20 2D Cartesian grid; each node has 4 actions (excepting edge nodes). b) Cosine similarity among the 200 node states $p$

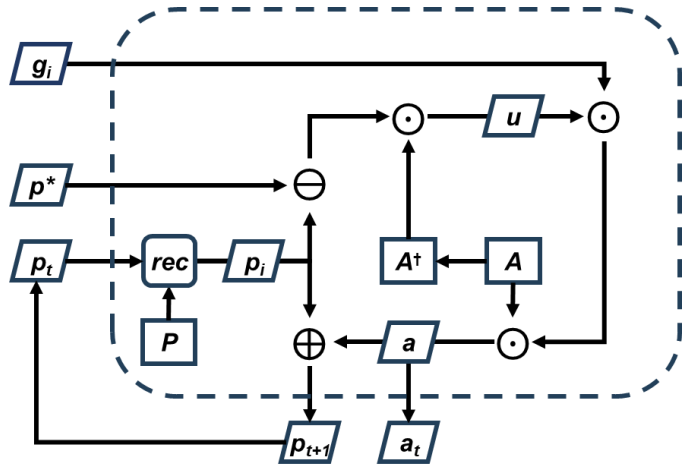

Figure 5. Block diagram of modified grid position CML $C_P$, replacing node specific gating matrix $G$ with touch sensor inputs $g$.

Here particularly, the *map* was calculated as,

$$map = sgn([sgn(a \odot p_1) + sgn(b \odot p_2) + sgn(c \odot p_3) + sgn(d \odot p_4) + sgn(e \odot p_5) \\ + sgn(k \odot p_6) + sgn(t \odot p_7) + sgn(h \odot p_8) + \eta]), \quad (17)$$

where $p_i$ corresponded to each of the 8 (out of 200) grid node positions used per trial and $\eta$ ensured the final *map* values remained bipolar. Recall, no matter how many *object* $\odot$ *position* pairs are added together, the *map* hypervector remains of length $d$. As will be explained in Section 4, the eight grid positions of these eight objects were also stored in an explicit memory matrix, $P_O$.

During operation, the robot queried the *map* for the target grid position $p^*$ of the next goal object prescribed by the *policy*,

$$p^* = \rho_{-i}(policy) \odot map. \quad (18)$$

For example, the first policy goal $k$,

$$p^* = \rho_{-1} \,(policy) \odot map$$
$$= {\sim}k \odot [k \odot p_6 + \eta\,]$$
$$\sim p_6. \qquad (19)$$

At this point, one could also update the gating matrix $G$ of the object CML to encode additional information such as the Euclidean distance between object positions in each maze, biasing the CML towards shorter physical lengths on the grid. In practice though, working with weighted graphs tended to require removing previously traversed edges to prevent dithering between nodes; so only unweighted edges were considered for these experiments.

### 3.5 Modular, hierarchical ML framework for sequential goal completion

Thus far, the robot has a) an object CML $C_O$ to store the relative location of maze objects; b) a grid position CML $C_P$ to navigate an open 2D grid space; and c) a *map* of objects and their grid positions within the maze. With all these pieces now available, the robot executed the *policy* on an arbitrary maze (Fig. 6, Algorithm A.2). 1) The calculated *policy*, Eq. 13), was provided as an input. 2) The *policy* was unpermuted once, which produced an approximate version of the target object state, $o^* = \rho_{-t}(policy)$. 3) The object CML iteratively plotted a path to this target state, $C_O(o^*, o_t)$. 4) The object CML returned the next planned subgoal object $\hat{o}_{t+1}$ approaching $o^*$. 5) The grid position of the subgoal object $\hat{o}_{t+1}$ was approximated from the *map*, $\hat{o}_{t+1} \odot map \sim p_o$. 6) As will be described later, preprocessing recovery step over the 8 known positions $P_O$ was more effective than a recovery step over all 200 grid position states, $rec(p_o, P_O) = p^*$. 7) The grid position CML recursively plotted a path to this target position, $C_P(p^*, p_t)$. 8) Each new grid position $p_t$ was 9) queried of the *map* to ascertain if any object existed at that location. 10) Until the robot reached $p^*$, the result $p_t \odot map \sim \eta$ was ignored by the object CML. When the object CML reached the target object $o^*$ from the *policy*, 1) the *policy* was unpermuted and process began again.

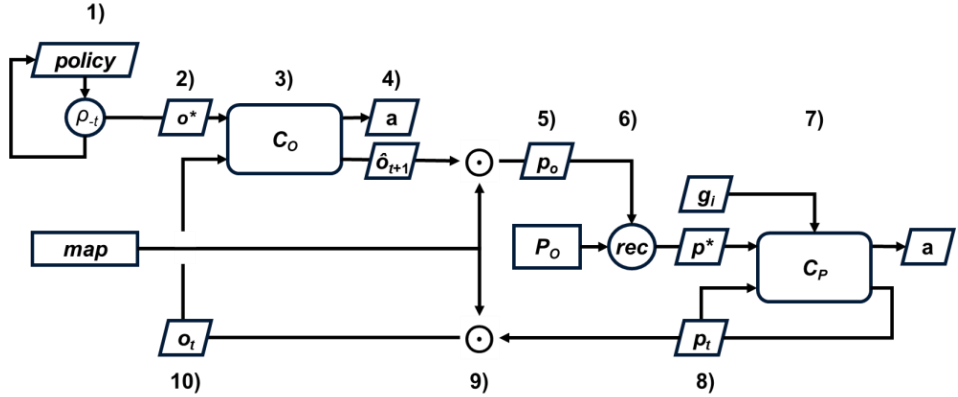

Figure 6. Wiring diagram for sequential goal completion task.

## 4. RESULTS

### 4.1 Grid position CML analysis

For an abstract graph CML, a separate pseudo-inverse of the action matrix, $A^\dagger$, was required for calculating the utility of each of edge action, (Eq. 5), Fig. 3b). In contrast, the regularity of grid node state hypervectors $P$ permitted simply using the transpose of $A$ instead,

$$u = A^\mathsf{T} (s^* - s_t). \qquad (20)$$

However, these grid node states included multiple duplicate vectors, precluding any reliable recovery over the full dictionary, $rec(p_t, P)$. Instead, a smaller dictionary of the eight object positions, $P_{1-8} = [p_1, \ldots, p_8]$, was created; and the recovery operation was performed over this dictionary, $rec(p_t, P_{1-8})$. The second challenge with these parallel and anti-parallel node state similarities was that not all possible grid positions produced maps which all eight object positions could be unambiguously recovered, that is, the recovery operator returned the incorrect grid position vector. If a *map* failed this check, a new random maze had to be generated. For 100 arrangements of objects in 20 mazes each, a viable maze was generated only $0.21 \pm 0.03$ of the time. Reported results below are only for these viable mazes.

The grid position CML was not trained with any obstructions, so to enable it to interact with the walls of the maze, touch sensors were provided. When calculating the state update, Eq. 6), the Winner Take All (WTA) function picked the action with the largest utility value; however, there were cases where the available actions moved the robot away from its target position, resulting in negative action utility values. Since invalid actions had a gated utility value of 0, the WTA selected an illegal action in those cases. An additional computation step was required to select the largest non-zero score, even if negative.

Unsurprisingly, when the robot was tasked to traverse from the key to the treasure using only the grid position CML and touch sensors, the robot succeeded only a fraction of the time, $0.43 \pm 0.09$ in 100 trials, typically due to dithering between two grid positions. For example, in Fig. 7 the robot had no means of discovering that there was a door further south. An additional path tracker to inhibit retraversing grid points would be required to mitigate this failure case. The objects in the object CML therefore served as anchor points, or known waypoints, over the maze. By integrating the two CMLs together via the *map*, the robot was able to access different position resolutions (actual vs. relative) for improved path planning.

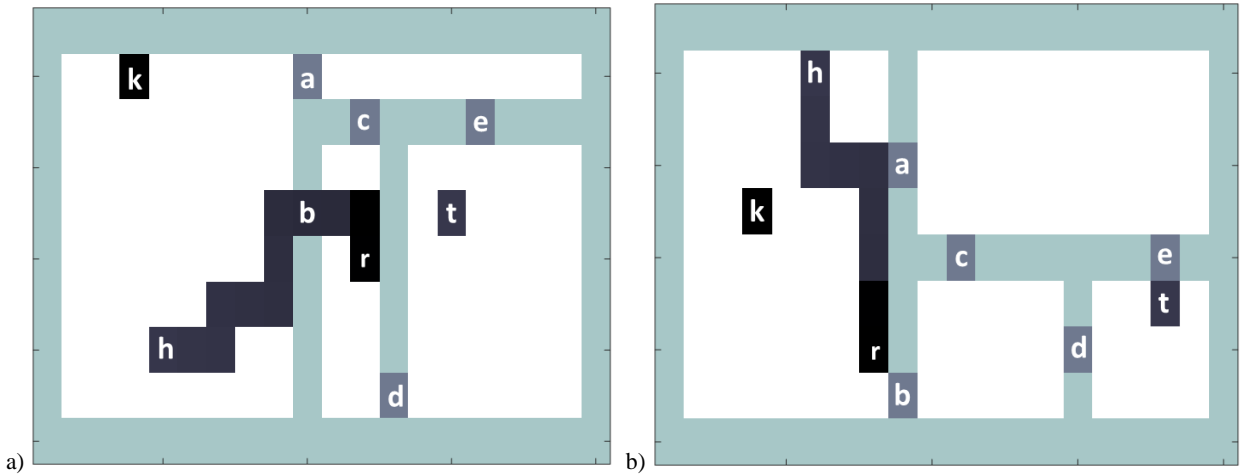

Figure 7. Since the robot *r* cannot find a) door *d* or b) door *b* below it, it dithers between two positions along the wall.

### 4.2 Sequential goal completion

The user-specified *policy* (Eq. 13) tasked the robot to navigate from home *h* 1) to the key *k*, then 2) to the treasure *t*, then 3) back to home *h*. When this *policy* was queried of the object CML only, it returned 1) a path of $\{h, k\}$ for the first task, then 2) either $\{k, a, e, t\}$ or $\{k, b, d, t\}$ for the second task, and finally 3) $\{t, e, a, h\}$ or $\{t, d, b, h\}$. By integrating the object CML with the grid CML for robot path planning and navigation, a simulated robot autonomously successfully navigated to each of these objects in the *policy* in all 50 different mazes (Fig. 8). Note, while goal 3) merely reversed direction of the object path, the grid CML often plotted a different grid path, illustrating that the CML did not memorize a fixed path between nodes.

While the object CML included all line-of-sight paths, during deployment, it was possible to modify object accessibility. For example, a random door was permanently closed, no longer permitting access to objects on the other side. Such a change was implemented in the CML-HDC framework simply by editing the gating matrix $G$ in the object CML $C_O$ to set the weights of all edge actions to/from the inaccessible door node to zero. No further updates were required in the rest of framework. For 50 trials, a single door was randomly removed (editing $G$ in $C_O$), and CML-HDC system successfully rerouted around the locked door in all trials (Fig. 9).

### 5. DISCUSSION

The described modular, hierarchical ML framework for sequential goal completion provides a neuroplausible approach to encoding both instincts and local environmental particulars. Certain activities, such as walking, are foundational; and, once learned, should not be easily modified. Here, the grid CML learned a 2D physical space and no further modifications were ever made, though there is no inherent limitation to the size of the encoded grid space beyond physical storage limitations. Instead, an object CML was created to track encountered objects. Spatial or semantic relationships can be dynamically encoded as edge weights. Further, the use of the *map* hypervector allowed both independently created CMLs to interact, serving as a short-term memory repository of anchor points. Thus, the grid CML and touch sensor array provided fine grained navigation through the maze, while the object CML provided anchor points for the robot to deliberately navigate towards. Changing the set of goals or set of available objects was restricted to precise locations in the CML-HDC framework, requiring no global retraining. To change the sequence of goals, one changed the *policy* hypervector. To change the position of objects with respect to the 2D grid, one updated that entry of the *map* hypervector. To add new objects to the object CML would require updates to all three ANNs: node state $O$, edge action $A$, and gating $G$. But, aside from possible indexing changes in the gating matrix, the new state and action vectors do not alter any previously created vectors in these matrices.

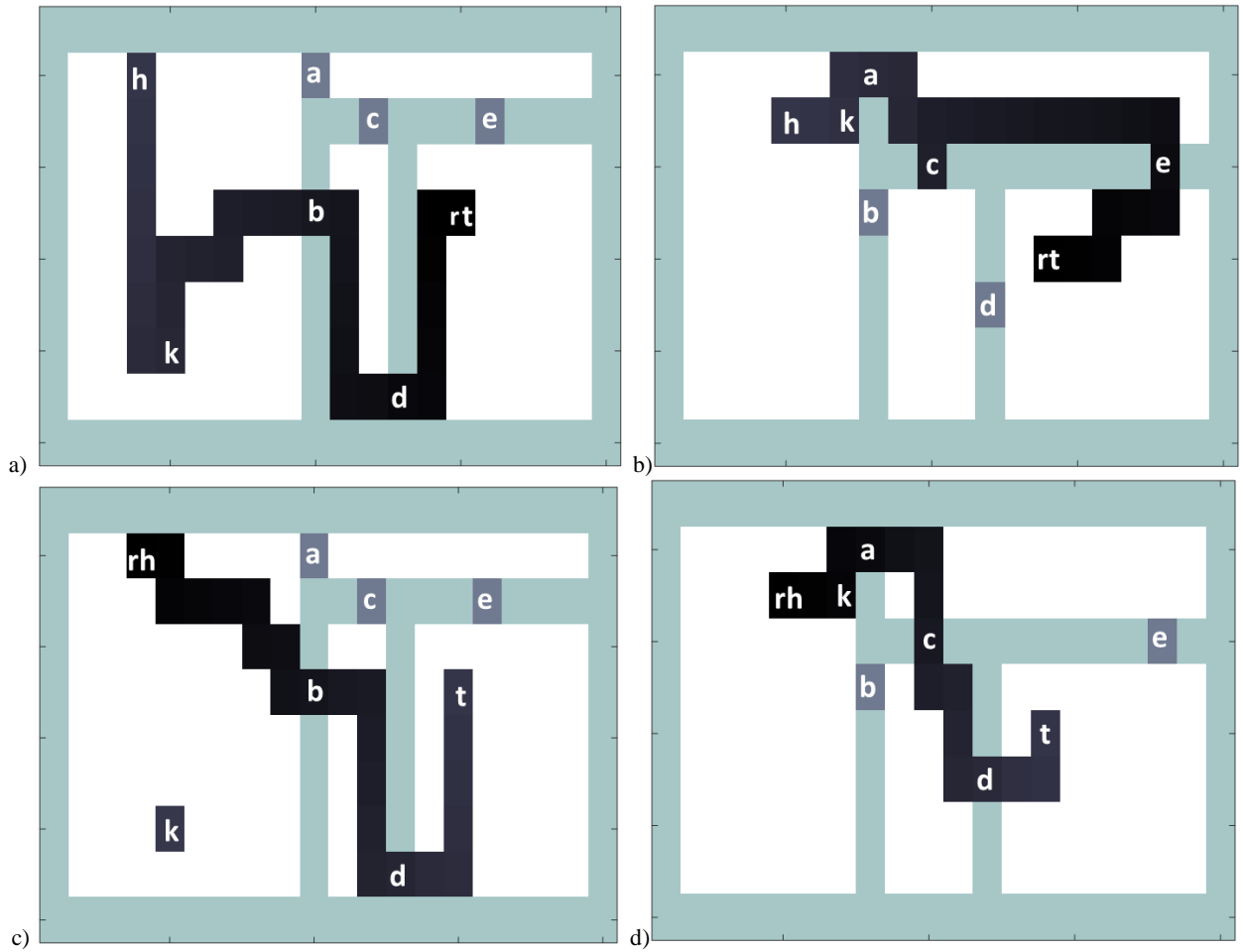

Figure 8. Paths generated for the robot *r* moving a) & b) from home *h* to key *k* to treasure *t* and then c) & d) from treasure *t* to home *h*. Paths a) & c) traverse same set of objects both ways, though along different paths; whereas, b) & d) take different routes entirely.

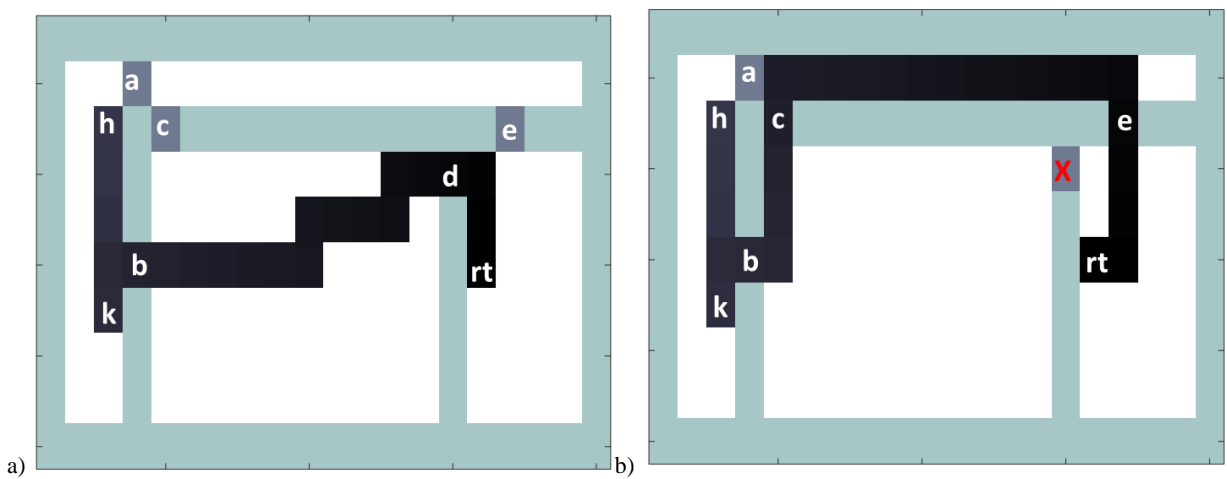

Figure 9. Planned robot *r* path from home *h* to the treasure *t* a) with all doors and b) after removing door d (red X.)

A similar maze-solving robot via HDC solution described the problem as behavior prioritization, where the robot may have to decide to move away from randomly distributed blocks/walls (based on touch sensors) to eventually move closer to the target position (based on goal position sensors) [12]. This robot had 7 sensors, viz. 4 touch sensors, 2 target displacement sensors, and 1 momentum (prior movement) sensor; and the paper evaluated several wiring diagrams using 2 addition and 1 multiplication operator to combine different sets of sensor information together to allow the robot to solve an arbitrary maze for a single target. The final HDC behavioral policy solution in effect became the sum of the exhaustive enumeration of all possible sensor-action combinations, attaining nearly 90% of the randomly generated mazes. While their behavioral policy was significantly simpler than Fig. 6, the simulated robot was purely reactionary and without a map, so failure cases were consistent with the dithering previously mentioned when just using the grid position CML (Fig. 7) and required a separate position tracker to prevent backtracking.

The more structured maze used in this work was inspired by the mazes and sequential tasks found in the reinforcement learning programs Minigrid [13] and MetaArcade [14]. While no reinforcement learning was implemented in this work, it is supported by the CML learning rules; so future work will seek to solve the full range of mazes and games in these environment suites. Additionally, the use of directed graphs might be considered to encode concepts of causality, e.g. a key unlocks a locked door, and correlation, e.g. the red key unlocks only the red door.

This work focused on bipolar vectors for HDC computation and real-valued vectors for the CMLs. However, complex-valued high dimensional vectors may also be used for both. In particular, a phasor (or phase vector) is a complex number corresponding to a spike time with respect to a local oscillator, $e^{i2\pi(t/r)}$, where $t$ is the integer time step with respect to the period resolution $r$. There are several recent papers illustrating spiking neural networks (SNN) based on resonate and fire neurons [15] and methods for implementing HDC operations directly in this SNN framework [16, 17]. While this work focused on path planning not simultaneous localization and mapping (SLAM) algorithms directly, [18] did implement a SLAM model using complex-valued HDC for SNN called spatial sematic pointers (SSP). As such, their learned hypervectors may be then used directly (or with slight modifications) to construct the CMLs described here for subsequent path planning.

Lastly, this work arbitrarily generated object hypervectors, but ideally there would be a neural network-based mechanism for consistently generating hypervector symbols from real-world sensor data. By working with task-agnostic feature extractors, e.g. CLIP [19] or DINO [20], one can translate raw data into semantically meaningful hypervectors. DNNs become no longer monolithic solutions but actually another ML module: an ML "analog to digital convertor" (A2D). For example, the Constrained-Few Shot Class Incremental Learning (C-FSCIL) framework used a pre-trained (and frozen) ResNet-12 feature extractor to populate a dictionary of hypervectors [21]. Lastly, the use of explicit knowledge dictionaries anticipates collaborative learning, where multiple ML agents can learn diverse things, yet because they share similar semantic dictionaries, they may collaboratively build and share knowledge graphs [22].

## 6. CONCLUSION

Sequential goal solving via deep neural networks (DNN) is a challenging task, especially when solutions must account for multiple competing subgoals. This work describes a modular, hierarchical machine learning (ML) framework integrating two emerging ML techniques: 1) cognitive map learners (CML) and 2) hyperdimensional computing (HDC), to sequentially navigate through a user-specified sequence of objects distributed in a variable maze. CMLs are used for path planning in abstract graphs, while HDC is an ML algebra based on high-dimensional vectors. By encoding each CML node state as a high-dimensional vectors multiple, independently trained CMLs were assembled together via HDC to navigate a maze to solve a sequential goal task. Critically, changes to goals or the sequence of goals only incurred localized changes in the CML-HDC framework, as opposed to a global DNN retraining schemes. This framework therefore enabled a more traditional engineering approach to ML solutions, with interpedently optimizable and arbitrarily configurable components.

## ACKNOWLEDGEMENTS

Any opinions, findings and conclusions, or recommendations expressed in this material are those of the authors, and do not necessarily reflect the views of the US Government, the Department of Defense, or the Air Force Research Lab. Distribution Statement A. Approved for Public Release; Distribution Unlimited: AFRL-2024-1629.

## REFERENCES

[1] N. Sunderhauf, O. Brock, W. Scheirer, et al., "The limits and potentials of deep learning for robotics," *International Journal of Robotics Research*, vol. 37, no. 4-5, pp. 405–420, 2018

[2] S. Russell and P. Norvig, *Artificial Intelligence: A modern approach*, Hoboken, NJ: Pearson, 2021

[3] C. Stöckl, Y. Yang, and W. Maass, "Local prediction-learning in high-dimensional spaces enables neural networks to plan," *Nature Communications* 15, no. 1, 2344, 2024

[4] D. Kleyko, D. Rachkovskij, E. Osipov, A. Rahimi, "A Survey on Hyperdimensional Computing aka Vector Symbolic Architectures, Part I: Models and Data Transformations," *ACM Computing Surveys*, vol. 55, is. 6, no. 130, pp 1–40, 2023

[5] D. Kleyko, D. Rachkovskij, E. Osipov, A. Rahimi, "A Survey on Hyperdimensional Computing aka Vector Symbolic Architectures, Part II: Applications, Cognitive Models, and Challenges," *ACM Computing Surveys*, vol. 55, is. 9, no. 175, pp 1–52, 2023

[6] P. Kanerva, "Hyperdimensional computing: An introduction to computing in distributed representation with high-dimensional random vectors," *Cognitive computation*, vol. 1, pp.139-159, 2009

[7] R. Gayler, "Multiplicative binding, representation operators & analogy (workshop poster)," 1998

[8] T. Plate, "Holographic reduced representations," *IEEE Transactions on Neural networks*, vol. 6, no. 3, pp. 623-641, 1995

[9] P. Neubert, S. Schubert, and P. Protzel, "An introduction to hyperdimensional computing for robotics," *KI-Künstliche Intelligenz*, vol. 33, pp. 319-330, 2019

[10] J. Hertz, A. Krogh, and R. Palmer, *Introduction to the theory of neural computation*. CRC Press, 2018

[11] McDonald, Nathan. "Modularizing and assembling cognitive map learners via hyperdimensional computing." *arXiv preprint arXiv:2304.04734*, 2023

[12] A. Menon, A. Natarajan, L. Olascoaga, Y. Kim, B. Benedict, and J. Rabaey, "On the role of hyperdimensional computing for behavioral prioritization in reactive robot navigation tasks," In *2022 International Conference on Robotics and Automation (ICRA)*, pp. 7335-7341. IEEE, 2022

[13] M. Chevalier-Boisvert, B. Dai, M. Towers, R. Perez-Vicente, L. Willems, S. Lahlou, S. Pal, P. Castro, and J. Terry. "Minigrid & miniworld: Modular & customizable reinforcement learning environments for goal-oriented tasks," *Advances in Neural Information Processing Systems* 36, 2024

[14] E. Staley, C. Ashcraft, B. Stoler, J. Markowitz, G. Vallabha, C. Ratto, and K. Katyal, "Meta arcade: A configurable environment suite for deep reinforcement learning and meta-learning," *Deep RL Workshop NeurIPS 2021,* 2021

[15] E. Frady, and F. Sommer, "Robust computation with rhythmic spike patterns," *Proceedings of the National Academy of Sciences* 116, no. 36, 18050-18059, 2019

[16] J. Orchard, and R. Jarvis. "Hyperdimensional Computing with Spiking-Phasor Neurons," In *Proceedings of the 2023 International Conference on Neuromorphic Systems*, pp. 1-7, 2023

[17] N. McDonald, L. Loomis, R. Davis, J. Salerno, A. Stephen, and C. Thiem, "Integrating complex valued hyperdimensional computing with modular artificial neural networks," In *Disruptive Technologies in Information Sciences VII*, vol. 12542, pp. 152-170. SPIE, 2023

[18] N. Dumont, P. Michael Furlong, J. Orchard, and C. Eliasmith. "Exploiting semantic information in a spiking neural SLAM system." *Frontiers in Neuroscience* 17, 1190515, 2023

[19] A. Radford, J. Kim, C. Hallacy, A. Ramesh, G. Goh, S. Agarwal, G. Sastry, et al., "Learning transferable visual models from natural language supervision," In *International conference on machine learning*, pp. 8748-8763. PMLR, 2021

[20] M. Oquab, T. Darcet, T. Moutakanni, H. Vo, M. Szafraniec, V. Khalidov, P. Fernandez, et al., "Dinov2: Learning robust visual features without supervision" *arXiv preprint arXiv:2304.07193*, 2023

[21] M. Hersche, G. Karunaratne, G. Cherubini, L. Benini, A. Sebastian, and A. Rahimi, "Constrained few-shot class-incremental learning," IEEE/CVF Conference on Computer Vision and Pattern Recognition, pp. 9057-9067, 2022

[22] P. Poduval, H. Alimohamadi, and M. Imani. "Graphd: Graph-based hyperdimensional memorization for brain-like cognitive learning." *Frontiers in Neuroscience* 16, 757125, 2022

# APPENDIX

TABLE I. KEY SYMBOLS AND DEFINITIONS

| Category | Symbol | Definition |
|---|---|---|
| CML | $s \in S$ | node state |
| | $a \in A$ | edge action |
| | $g \in G$ | edge action availability |
| | $s_t$ | current node state |
| | $\hat{s}$ | predicted/prescribed node state |
| | $s^*$ | target node state |
| | $u$ | action utility |
| HDC | $d$ | hypervector length |
| | $\eta$ | random hypervector |
| | $\delta$ | cosine similarity |
| | sgn | sign operator |
| | $\odot$ | elementwise multiplication |
| | rec | recovery |
| | $\theta$ | similarity threshold/noise floor |
| | $\rho$ | permutation |
| | $\phi$ | similarity threshold when $s^* \approx s_t$ |
| Maze | $h$ | home state |
| | $k$ | key state |
| | $t$ | treasure state |
| | $a,b,c,d,e$ | door states |
| | $o \in O$ | object node state |
| | $C_O$ | object CML |
| | $p \in P$ | grid position node state |
| | $C_P$ | grid position CML |
| | $p_{1\text{-}8} \in P_O$ | subset of 8 object positions |

---

**Algorithm A.1:** $C(s^*, s_t)$ with feedback loop

---

**comment** Plan path from node state $i$ to $j$, with feedback from $\hat{s}_{t+1}$ to $s_t$

**if** $\delta(s^*, S) \geq \theta$      % if $s^*$ is a state in $S$

     $s_j \leftarrow rec(s^*, S)$      % recover target state $s_j$

     $s_i \leftarrow rec(s_t, S)$      % recover current state $s_i$

     **while** $\delta(s_j, s_i) < \phi$      % while $s_i$ is insufficiently similar to $s_j$

         $d \leftarrow s_j - s_i$      % "direction" of target

         $u \leftarrow A^\dagger d$      % utility score per edge action in $A$

         $g_i \leftarrow G(:, i)$      % gating vector is $i^{th}$ column in $G$

         $a \leftarrow A \cdot WTA(g_i \odot u)$      % *WTA* algorithm selects single action matrix column \

         $\hat{s}_{t+1} \leftarrow s_i + a$      % calculate next state as sum of current state and action

     **end**

     **return** $a, \hat{s}_{t+1}$

**else**      % else if $s^*$ is not a state in $S$

     **return** 0

**end**

**Algorithm A.2:** Sequential goal completion

**comment** Navigate to all objects in *policy*

**while** $\delta(policy, O) \geq \theta_O$   % while *policy* is an object in $O$
   $policy \leftarrow \rho_{-1}(policy)$   % unpermute *policy*
   $o^* \leftarrow policy$   % *policy* is target object
   **while** $\delta(o^*, o_t) < \phi_O$   % while current object $o_t$ is insufficiently similar to target object $o^*$
     $\hat{o}_{t+1} \leftarrow C_O(o^*, o_t)$   % plan subgoal object state $\hat{o}_{t+1}$ approaching $o^*$
     $p_o \leftarrow map \odot \hat{o}_{t+1}$   % approximate position $p_o$ of subgoal object $\hat{o}_{t+1}$
     $p^* \leftarrow rec(p_o, P_{1\text{-}8})$   % recover object position
     **while** $\delta(p^*, p_t) < \phi_G$   % while current grid position $p_t$ insufficiently similar to $p^*$
       $p_t \leftarrow C_P(p^*, p_t)$   % plan next grid position $p_t$ approaching $p^*$
     **end**
   **end**
   $o_t \leftarrow map \odot p_t$   % retrieve object $o_t$ at position $p_t$
**end**